\documentclass{article}
\usepackage{graphicx}

\usepackage[preprint]{spconf}

\usepackage{amsmath}
\usepackage{amssymb}
\usepackage{bm}
\usepackage{url}
\usepackage{stmaryrd}

\usepackage{pifont}
\newcommand{\xmarksimple}{\ding{56}}%
\newcommand{\starsimple}{\ding{51}}%

\usepackage{xcolor}
\usepackage{colortbl}
\usepackage{multirow}

\usepackage{CJKutf8}
\newcommand*{\jpn}[1]{\begin{CJK}{UTF8}{ipxm}#1\end{CJK}}

\newcommand{\red}[1]{\textcolor{red}{#1}}
\newcommand{\blue}[1]{\textcolor{blue}{#1}}
\newcommand{\age}  {{\normalfont\red{[}}}
\newcommand{\sage} {{\normalfont\blue{]}}}

\def\vsp{~~}
\def\NN{\textsc{n}}
\def\QQ{\textsc{q}}

\usepackage[normalem]{ulem}

\title{
    Accent Estimation of Japanese Words From Their Surfaces and Romanizations
    For Building Large Vocabulary Accent Dictionaries
}

\name{Hideyuki Tachibana \qquad Yotaro Katayama
\thanks{
    This paper is based on results obtained from a project subsidized by
    the New Energy and Industrial Technology Development Organization (NEDO).
}%
}
\address{PKSHA Technology Inc., Hongo, Bunkyo, Tokyo, Japan}
\begin{document}

\noindent
    \copyright\ IEEE 2020. 
    Personal use of this material is permitted. 
    Permission from IEEE must be obtained for all other uses, 
    in any current or future media, 
    including reprinting/republishing this material for advertising or promotional purposes, 
    creating new collective works, 
    for resale or redistribution to servers or lists, 
    or reuse of any copyrighted component of this work in other works.
\newpage

\ninept
\maketitle
\copyrightnotice{\copyright\ IEEE 2020}
\toappear{Published as a conference paper at {\it ICASSP 2020}}
\begin{abstract}
    In Japanese text-to-speech (TTS),
    it is necessary to add accent information
    to the input sentence.
    However, there are a limited number of publicly available accent dictionaries,
    and those dictionaries e.g.{} UniDic, do not contain many compound words, proper nouns, etc.,
    which are required in a practical TTS system.
    In order to build a large scale accent dictionary
    that contains those words,
    the authors developed an accent estimation technique that
    predicts the accent of a word from its limited information,
    namely the surface (e.g. kanji) and the yomi (simplified phonetic information).
    It is experimentally shown that the technique can estimate accents with high accuracies,
    especially for some categories of words.
    The authors applied this technique to an existing large vocabulary Japanese dictionary NEologd,
    and obtained a large vocabulary Japanese accent dictionary.
    Many cases have been observed in which the use of this dictionary yields more appropriate phonetic information than UniDic.
\end{abstract}

\section{Introduction}\label{sec:intro}
\begin{figure}[!t]
        \begin{minipage}[t]{0.99\linewidth}
              \centering
              \centerline{\includegraphics[width=0.75\columnwidth]{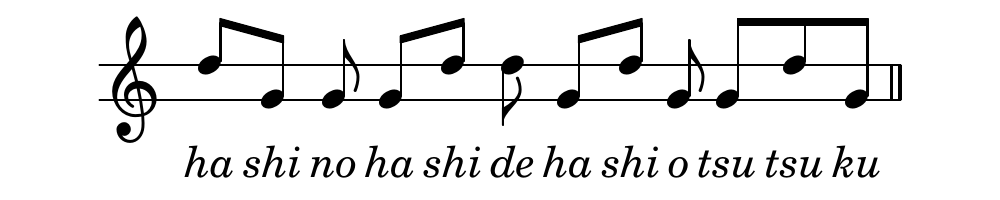}}
              \vspace{-10pt}
              \caption{
                Concept of Japanese accent. Each note indicates a mora.
              }\label{fig:hashiga}
        \end{minipage}
    \end{figure}

    Japanese text is composed of variety of characters,
    and each character is pronounced in various ways depending on the context.
    Therefore, the first task of Japanese TTS is to convert the raw text into some phonetic information as follows,
    using some dictionaries.
    \begin{align*}
        \text{raw text} &: \text{\jpn{箸の端で橋をつつく。}} \\
        \text{yomi} &: \text{\textit{hashi no hashi de hashi o tsutsuku.}}
    \end{align*}
    However, the standard Hepburn romanization, which we call `\textit{yomi}'\footnote{
        Yomis are often written in \textit{kana} characters (\textit{hiragana} or \textit{katakana}),
        but we show them using Latin letters (\textit{romaji}) in this paper for readability.
        Kana and romaji are essentially almost the same.
    }
    in this paper, is not sufficient yet,
    as it lacks of the accent information of each word,
    which sometimes even changes the meaning of it
    (see Table~\ref{fig:jpnwords}.)
    Therefore, we need to insert appropriate accent marks as follows,
    \[
        \text{phonetic}: \text{{ha\sage{}shi no ha\age{}shi de ha\age{}shi\sage{} o tsu\age{}tsu\sage{}ku.}}
    \]
    where the brackets ``\age'' and ``\sage'' indicate ``raise the pitch'' and ``lower the pitch,'' respectively\footnote{
        It has been common in Japanese TTS systems to use the binary pitch model
        that the pitch of a mora is either H (high) or L (low).
        However, some linguists claim that the model based on \age{} and \sage{} is closer to the actual speech.
        See e.g.{} Uwano's articles~\cite{asakura,uwano2009two}.
    }.
    Intuitively, it is pronounced like the `melody' shown in Fig.~\ref{fig:hashiga}.

    \begin{table}[t]
        \begin{minipage}[t]{0.95\linewidth}
            \caption{
                Examples of Japanese words
                whose meaning depend on the accents in Tokyo dialect.
            }\label{fig:jpnwords}
                {\scriptsize
                \begin{tabular}{@{}p{0.12\columnwidth}p{0.16\columnwidth}|p{0.24\columnwidth}p{0.36\columnwidth}@{}}
                    \hline
                    surface & yomi & accent (Tokyo)& meaning \\ \hline
                        \jpn{酒} & \textit{sake} & sa\age ke & alcoholic beverage\\
                        \jpn{鮭} & \textit{sake} & sa\sage ke & salmon \\
                        \jpn{藤} & \textit{fuji} & fu\age ji & wisteria\\
                        \jpn{富士} & \textit{fuji} & fu\sage ji & Mt.{} Fuji \\
                        \jpn{玉} & \textit{tama} & ta\age ma\sage & ball \\
                        \jpn{多摩} & \textit{tama} & ta\sage ma & Western Tokyo \\
                        \jpn{伝記} & \textit{denki} & de\age \NN{}ki & biography \\
                        \jpn{電気} & \textit{denki} & de\sage \NN{}ki & electricity \\\hline
                \end{tabular}
                }
        \end{minipage}
        \vspace{-15pt}
    \end{table}
    \begin{table}[t]
        \begin{minipage}[t]{0.99\linewidth}
            \caption{
                Examples of words that MeCab+UniDic does not analyze correctly. 
            }\label{fig:jpnwrongyomi}
                {\scriptsize
                \begin{tabular}{@{}p{0.21\columnwidth}p{0.30\columnwidth}p{0.4\columnwidth}@{}}
                    \hline
                    surface & correct yomi & wrong yomi based on UniDic \\ \hline
                    \jpn{一日千秋} & \textit{ichijitsusensh\^u} & \textit{ichi\vsp{}nichi\vsp{}chiaki} \\
                    \jpn{御御御付け} & \textit{omiotsuke} & \textit{go\vsp{}go\vsp{}go\vsp{}tsuke} \\
                    \jpn{IEEE} & \textit{aitoripuruii} & \textit{ai\vsp{}ii\vsp{}ii\vsp{}ii} \\
                    \jpn{36協定} & \textit{saburoku\vsp{}ky\^ot\^e} & \textit{san\vsp{}roku\vsp{}ky\^ot\^e} \\
                    \jpn{山東京伝} & \textit{sant\^o\vsp{}ky\^oden} & \textit{yama\vsp{}t\^oky\^o\vsp{}den} \\
                    \jpn{県犬養三千代} & \textit{agatanoinukainomichiyo} & \textit{ken\vsp{}inukai\vsp{}michiyo} \\
                    \jpn{諸葛亮孔明} & \textit{shokatsury\^ok\^om\^e} & \textit{shokatsu\vsp{}ry\^o hiroaki} \\
                    \jpn{八幡山} & \textit{hachiman'yama} & \textit{yawata\vsp{}yama} \\
                    \jpn{本八幡} & \textit{motoyawata} & \textit{hon\vsp{}hachiman} \\
                    \jpn{武蔵嵐山} & \textit{musashiranzan} & \textit{musashi\vsp{}arashiyama} \\
                    \jpn{嶺上開花} & \textit{rinshankaih\^o} & \textit{r\^e\vsp{}j\^o\vsp{}kaika} \\ \hline
                \end{tabular}
                }
        \end{minipage}
    \end{table}

    Since the accent marks are not explicitly written in the raw text nor the yomi,
    we need to look them up in some dictionaries,
    but the number of accent dictionaries publicly available is limited.
    At the moment, UniDic~\cite{unidiceng,unidic,unidicman},
    an open source Japanese dictionary for a text analyzer MeCab~\cite{mecab},
    is one of the few options,
    but it has a shortcoming that many words are intentionally excluded,
    e.g., compound words, proper nouns, idiomatic phrases, numerals, technical terms, etc.
    because of its policy to prioritize the linguistic consistency.
    As the cost of that,
    it often fails to give the correct yomis to some compound words and proper nouns e.g.{} shown in Table~\ref{fig:jpnwrongyomi}.
    Thus we need a dictionary that contains those words and their correct yomis and accents.

    The objective of this paper is to propose a technique for
    building a large scale Japanese accent dictionary that covers such words,
    using limited information of them,
    viz.{} their surfaces and yomis.
    Fortunately, there already exists NEologd~\cite{sato2016NEologdipsjnl229,%
        sato2017mecabipadicNEologdnlp2017}\footnote{
            \url{https://github.com/neologd/mecab-unidic-neologd}.
            We used 27/Dec/2018 version in this paper.},
    a web crawling-based large scale dictionary for MeCab,
    which is recently very popular in Japanese NLP.
    The dictionary contains millions of pairs of surfaces and yomis,
    and thus, we may construct a large scale accent dictionary
    just by applying our technique to it.
    To our knowledge, there has not been such a large vocabulary Japanese accent dictionary
    whose vocabulary size is as large as several millions.

\section{Related Work}
    In both linguistics and engineering,
    there have been many studies on accent of Japanese.
    Of these, Sagisaka's rule~\cite{sagisaka} would be a well-established classic in the engineering community,
    and it (and its complements and extensions e.g.{}~\cite{miyazaki,kita2002}) has been exploited in many Japanese TTS systems and related applications,
    e.g.{} GalateaTalk~\cite{galatea}, OpenJTalk, Orpheus~\cite{orpheus}, etc.
    A shortcoming of such rule-based approaches is that
    the users need to enter the grammatical information of the neologisms correctly
    when they are going to add them to a custom dictionary.

    In addition to those rule-based approaches, some statistical techniques are also proposed.
    For example,
    Nagano et al.{} proposed N-gram based technique~\cite{nagano},
    and Minematsu, Suzuki et al.{} proposed a technique based on  CRF~\cite{minematsu,suzuki}.
    The CRF-based technique implicitly assumes that a text analyzer can separate a sentence into morphemes correctly,
    which is not always the case, as shown in Table~\ref{fig:jpnwrongyomi}.

    Other machine learning-based techniques include Bruguier's~\cite{bruguier2018sequence} method based on LSTM and an attention mechanism.
    The objective of the study is to construct an accent dictionary but the input data are different from ours;
    it exploits audio data, as well as yomi.

    Comparing to those existing methods, the advantages of our method would be as follows:
    (1) The user is required to enter only the accessible information of the word,
        namely the surface and the yomi, when adding it to the custom dictionary.
    (2) Our technique
        could be robust against the errors of MeCab+UniDic, as it searches
        somewhat plausible morphology
        from several candidates exploiting both surface and yomi.
    (3) We `pre-render' the accents of the compound words, proper nouns, etc., as many as possible,
        and list them in the dictionary.
        Without postprocessing modules that estimate the accent sandhi,
        the dictionary alone gives the plausible accents of those words. This will make the system simpler.
    (4) We could exploit NEologd as a basis,
        which is a popular dictionary in the open-source ecosystem of Japanese NLP.
        Although further improvements are needed,
        we can obtain a very large scale accent dictionary at once.

\begin{table*}[!t]
    \centering{
        \begin{minipage}{0.95\textwidth}
            \centering
            {\scriptsize
                \caption{
                    (a)~Regex-based rough classification of the words listed in NEologd dictionary.
                    (The classification is not necessarilly correct.)
                    (b)~The number of words we annotated.
                    (c)~Result of our experiment on accent estimation.
                    Lighter color indicates better performance.
                }
                \label{tab:neologdwords}
                \hspace{-10pt}
                \par\smallskip\noindent
                \begin{tabular}{@{}
                        lrl
                        |r
                        |cccccc
                    }
                    \hline
                        \multicolumn{3}{c|}{ (a) } &
                        \multicolumn{1}{c|}{ (b) } &
                        \multicolumn{6}{c}{ (c) }
                        \\ \hline
                    & & &
                         &
                         & &
                         \multicolumn{2}{c}{raise \age{}} &
                         \multicolumn{2}{c}{lower \sage{}}
                        \\
                    category
                        & \#words
                        & example
                        & \#annot.
                        & EMR
                        & AHD
                        & prec.
                        & rec.
                        & prec.
                        & rec.
                        \\ \hline
                    ignored
                        & 326k
                        & \textit{noisy words}
                        & --
                        & --
                        & --
                        & --
                        & --
                        & --
                        & --
                        \\
                    emoji, symbol
                        & \leavevmode{\phantom{xx}4k}
                        & \jpn{(((o(*ﾟ▽ﾟ*)o)))}, $\spadesuit$, \jpn{♨}, \jpn{♪}, \jpn{㊗}
                        & 200
                        & \cellcolor{white!50!teal} 50\phantom{.0}\%
                        & \cellcolor{orange!27!white} 0.53
                        & \cellcolor{white!71!red} 0.71
                        & \cellcolor{white!89!red} 0.89
                        & \cellcolor{white!86!blue} 0.86
                        & \cellcolor{white!40!blue} 0.40
                        \\
                    company (K.K.)
                        & 193k
                        & \jpn{株式会社\textit{ XX }}, \jpn{（株）\textit{ XX }}
                        & 500
                        & \cellcolor{white!76!teal}  76\phantom{.0}\%
                        & \cellcolor{orange!27!white} 0.54
                        & \cellcolor{white!93!red} 0.93
                        & \cellcolor{white!93!red} 0.93
                        & \cellcolor{white!93!blue} 0.93
                        & \cellcolor{white!90!blue} 0.90
                        \\
                    company (Y.K.)
                        & \leavevmode{\phantom{x}13k}
                        & \jpn{有限会社\textit{ XX }}, \jpn{（有）\textit{ XX }}
                        & 500
                        & \cellcolor{white!78!teal}  78\phantom{.0}\%
                        & \cellcolor{orange!20!white} 0.39
                        & \cellcolor{white!97!red} 0.97
                        & \cellcolor{white!96!red} 0.96
                        & \cellcolor{white!96!blue} 0.96
                        & \cellcolor{white!90!blue} 0.90
                        \\
                    station
                        & \leavevmode{\phantom{x}24k}
                        & \jpn{\textit{ XX }駅}
                        & 500
                        & \cellcolor{white!82!teal}  82\phantom{.0}\%
                        & \cellcolor{orange!15!white} 0.31
                        & \cellcolor{white!91!red} 0.91
                        & \cellcolor{white!88!red} 0.88
                        & \cellcolor{white!94!blue} 0.94
                        & \cellcolor{white!90!blue} 0.90
                         \\
                    road
                        & \leavevmode{\phantom{x}12k}
                        & \jpn{\textit{ XX }県道\textit{ YY }号\textit{ ZZ }線}
                        & 500
                        & \cellcolor{white!44!teal}   44\phantom{.0}\%
                        & \cellcolor{orange!57!white} 1.15
                        & \cellcolor{white!93!red} 0.93
                        & \cellcolor{white!93!red} 0.93
                        & \cellcolor{white!92!blue} 0.92
                        & \cellcolor{white!87!blue} 0.87
                        \\
                    school
                        & \leavevmode{\phantom{x}28k}
                        & \jpn{\textit{ XX }県立\textit{ YY }高等学校}
                        & 500
                        & \cellcolor{white!81!teal}   81\phantom{.0}\%
                        & \cellcolor{orange!17!white} 0.35
                        & \cellcolor{white!93!red} 0.93
                        & \cellcolor{white!94!red} 0.94
                        & \cellcolor{white!96!blue} 0.96
                        & \cellcolor{white!93!blue} 0.93
                        \\
                    address
                        &  546k
                        & \jpn{\textit{ XX }県\textit{ YY }市\textit{ ZZ}}
                        & 1,000
                        & \cellcolor{white!56!teal}  56.0\%
                        & \cellcolor{orange!43!white} 0.86
                        & \cellcolor{white!92!red} 0.92
                        & \cellcolor{white!91!red} 0.91
                        & \cellcolor{white!89!blue} 0.89
                        & \cellcolor{white!81!blue} 0.81
                        \\
                    person (katakana)
                        & 382k
                        & \jpn{ポール・マッカートニー}
                        & 2,000
                        & \cellcolor{white!77!teal}  77.0\%
                        & \cellcolor{orange!19!white} 0.38
                        & \cellcolor{white!92!red} 0.92
                        & \cellcolor{white!95!red} 0.95
                        & \cellcolor{white!85!blue} 0.85
                        & \cellcolor{white!85!blue} 0.85
                        \\
                    person (kanji, kana)
                        & 549k
                        & \jpn{徳川家康}, \jpn{古今亭志ん生}
                        & 2,000
                        & \cellcolor{white!66!teal}  66.5\%
                        & \cellcolor{orange!20!white} 0.50
                        & \cellcolor{white!83!red} 0.83
                        & \cellcolor{white!89!red} 0.89
                        & \cellcolor{white!85!blue} 0.85
                        & \cellcolor{white!74!blue} 0.74
                        \\
                    person (other)
                        & \leavevmode{\phantom{x}98k}
                        & Smith, \jpn{「\textit{ XX }」製作委員会}
                        & 1,000
                        & \cellcolor{white!64!teal}  64.0\%
                        & \cellcolor{orange!37!white} 0.75
                        & \cellcolor{white!86!red} 0.86
                        & \cellcolor{white!85!red} 0.85
                        & \cellcolor{white!81!blue} 0.81
                        & \cellcolor{white!70!blue} 0.70
                        \\
                    numeral
                        & \phantom{x}88k
                        & 980.5hPa, 35kg, \$50
                        & 1,000
                        & \cellcolor{white!85!teal}  85.5\%
                        & \cellcolor{orange!12!white} 0.24
                        & \cellcolor{white!97!red} 0.97
                        & \cellcolor{white!98!red} 0.98
                        & \cellcolor{white!97!blue} 0.97
                        & \cellcolor{white!96!blue} 0.96
                        \\
                    date
                        & \phantom{xx}1k
                        & \jpn{10月21日}, \jpn{十月二十一日}
                        & 500
                        & \cellcolor{white!91!teal}  91\phantom{.0}\%
                        & \cellcolor{orange!9!white} 0.18
                        & \cellcolor{white!98!red} 0.98
                        & \cellcolor{white!97!red} 0.97
                        & \cellcolor{white!95!blue} 0.95
                        & \cellcolor{white!95!blue} 0.95
                        \\
                    numeral-like
                        & \phantom{xx}5k
                        & \jpn{100円ショップ, 3秒ルール}
                        & 500
                        & \cellcolor{white!56!teal}  56\phantom{.0}\%
                        & \cellcolor{orange!44!white} 0.88
                        & \cellcolor{white!90!red} 0.90
                        & \cellcolor{white!88!red} 0.88
                        & \cellcolor{white!87!blue} 0.87
                        & \cellcolor{white!80!blue} 0.80
                        \\
                    katakana words
                        & 263k
                        & \jpn{バスケットボールリーグ}
                        & 2,000
                        & \cellcolor{white!78!teal}  78.3\%
                        & \cellcolor{orange!16!white} 0.32
                        & \cellcolor{white!91!red} 0.91
                        & \cellcolor{white!96!red} 0.96
                        & \cellcolor{white!89!blue} 0.89
                        & \cellcolor{white!82!blue} 0.82
                        \\
                    romaji, some symbols(\texttt{'-!}, etc.)
                        & 125k
                        & Kubernetes, pink floyd
                        & 1,000
                        & \cellcolor{white!76!teal}  76.5\%
                        & \cellcolor{orange!19!white} 0.38
                        & \cellcolor{white!94!red} 0.94
                        & \cellcolor{white!91!red} 0.91
                        & \cellcolor{white!85!blue} 0.85
                        & \cellcolor{white!82!blue} 0.82
                        \\
                    kanji, kana
                        & 482k
                        & \jpn{類聚名義抄}, \jpn{可換環}, \jpn{こいぬ座}, \jpn{東京タワー}
                        & 2,000
                        & \cellcolor{white!53!teal}  53.5\%
                        & \cellcolor{orange!45!white} 0.91
                        & \cellcolor{white!83!red} 0.83
                        & \cellcolor{white!81!red} 0.81
                        & \cellcolor{white!76!blue} 0.76
                        & \cellcolor{white!66!blue} 0.66
                        \\
                    kanji, kana, romaji, some symbols
                        & \phantom{x}50k
                        & \jpn{Tシャツ, SDカード, W杯}
                        & 500
                        & \cellcolor{white!36!teal}  36\phantom{.0}\%
                        & \cellcolor{orange!72!white} 1.43
                        & \cellcolor{white!81!red} 0.81
                        & \cellcolor{white!75!red} 0.75
                        & \cellcolor{white!77!blue} 0.77
                        & \cellcolor{white!59!blue} 0.59
                        \\
                    other (remain)
                        & \phantom{x}81k
                        & \jpn{word2vec}, \jpn{1Q84}, \jpn{リスト::声優/あ行}
                        & 1,000
                        & \cellcolor{white!34!teal}  34.5\%
                        & \cellcolor{orange!91!white} 1.82
                        & \cellcolor{white!76!red} 0.76
                        & \cellcolor{white!70!red} 0.70
                        & \cellcolor{white!75!blue} 0.75
                        & \cellcolor{white!64!blue} 0.64
                        \\
                    \hline
                \end{tabular}
            }
        \end{minipage}
    }
\end{table*}

\section{Problem Definition}

    Let us assume that the surface $s$ and the yomi $y$ of a word are given.
    For example, $(s, y) = (\text{\jpn{深層学習}}, \textit{shins\^ogakush\^u})$ (meaning `deep learning').
    Note, using a simple subroutine, the yomi is mutually converted to a sequence of morae,
    so we also denote it as $y$, i.e.,
    \begin{equation}
        y = [\text{shi}, \text{N}, \text{so}, \text{o}, \text{ga}, \text{ku}, \text{shu}, \text{u}]^\mathsf{T}.
    \end{equation}
    Our target is the accent $\bm{a} \in \{+1, -1, 0\}^{|y|}$
    \begin{equation}
        \bm{a} = [~~~
        \text{\makebox[0pt]{$\stackrel{\large\text{shi}}{\phantom{\div}}$}}\makebox[2em]{$+1$,}
        \text{\makebox[0pt]{$\stackrel{\large\text{N}}{\phantom{\div}}$}}\makebox[2em]{$0$,}
        \text{\makebox[0pt]{$\stackrel{\large\text{so}}{\phantom{\div}}$}}\makebox[2em]{$0$,}
        \text{\makebox[0pt]{$\stackrel{\large\text{o}}{\phantom{\div}}$}}\makebox[2em]{$0$,}
        \text{\makebox[0pt]{$\stackrel{\large\text{ga}}{\phantom{\div}}$}}\makebox[2em]{$-1$,}
        \text{\makebox[0pt]{$\stackrel{\large\text{ku}}{\phantom{\div}}$}}\makebox[2em]{$0$,}
        \text{\makebox[0pt]{$\stackrel{\large\text{shu}}{\phantom{\div}}$}}\makebox[2em]{$0$,}
        \text{\makebox[0pt]{$\stackrel{\large\text{u}}{\phantom{\div}}$}}\makebox[2em]{$0$}
        \text{\makebox[0pt]{$\stackrel{\large\text{*}}{\phantom{\div}}$}}
        ~~~]^\mathsf{T},
    \end{equation}
    where $+1$ and $-1$ indicate \age{} and \sage{}, respectively.
    Our objective is to construct a function $f: (s, y) \mapsto \bm{a}$ using triples
    $\{(s^{(i)}, y^{(i)}, \bm{a}^{(i)})\}$.

    The problem setting is reasonable for the following two reasons.
    Firstly, let us consider a case where
    a native/fluent speaker is trying to add a newly-coined word (e.g.{} the name of their new product) to a custom dictionary.
    In this case, it may not be expected
    that they can enter neither the accent,
    (native/fluent speakers are not necessarily conscious of the accents of words),
    nor the grammatical information of the word
    i.e.{} POS tag, \textit{goshu}\footnote{
        Goshu indicates the origin of a word,
        i.e., whether a word is a Japanese word, a loanword from Chinese, or Western languages, etc.
    }, sandhi (liaison) rules, and accent sandhi type~\cite{unidicman,sagisaka}.
    However, we can expect that most native/fluent speakers
    at least know the surface and the yomi of the word they are going to add to their custom dictionary.
    Secondly, there already exists a large size dictionary publicly available, viz., NEologd,
    which contains approx 3 million pairs $\{(s^{(i)}, y^{(i)})\}_{1\le i \lesssim 3\times 10^6}$.

\section{Accent Estimation Technique}
    \subsection{Feature Extraction from Surface $s$}\label{sec:feature}
        Instead of using a raw $s$,
        we may extract detailed linguistic information from $s$ using MeCab+UniDic\footnote{
            As a preprocessing, we converted all the numerals in $s$
            into kanji~(for example, 10234.56 $\to$ \jpn{一万二百三十四点五六}) using a simple subroutine.
            \label{footnote:numeral}
        }.
        Let $\pi_r(s)$ be the $r$-th best result of MeCab+UniDic analysis.
        In general, $\pi_1(s)$ is not always the correct morphological segmentation of a compound word $s$.
        For example, by analyzing the word $s=$\jpn{一日千秋} ($y=$\textit{ichijitsusensh\^u}), we have
        \begin{equation*}
        \pi_1(s) =
            \left[\begin{array}{c|c|c}
            \text{\jpn{一} i\age chi\sage} & \text{\jpn{日} ni\sage chi} & \text{\jpn{千秋} chi\sage aki}\\
            \text{numeral, C3} &
            \text{suffix of numerals, C3} &
            \text{given name}
            \end{array}\right]
        \end{equation*}
        where ``C3'' is the accent sandhi type~\cite{unidicman} of the word.
        The $\pi_1(s)$ is not correct simply because the yomi is different from $y$.
        However, the $56$-th best result returns the correct yomi as follows,
        \begin{equation*}
        \pi_{56}(s) =
            \left[\begin{array}{c|c|c}
            \text{\jpn{一} i\age chi\sage} &\text{\jpn{日} jitsu} & \text{\jpn{千秋} se\age \NN{}shuu}\\
            \text{numeral, C3} &
            \text{suffix, C4} &
            \text{noun, C2}
            \end{array}\right]
        \end{equation*}

        In general,
        we may obtain a better morphological segmentation of a surface $s$
        by searching the $\pi_i(s)$ whose yomi is close to $y$.

        On the basis of this idea, we extracted $m$ candidates from the $20$-best analysis results $\{\pi_r(s)\}_{1\le r \le 20}$,
        based on Levenshtein distance from $y$,
        and sampled one $\pi^*(s)$ out of those $m$ candidates randomly for each iteration. ($m=3$ during training, and $m=1$ during inference.)
        From thus obtained $\pi^*(s)$,
        we extracted consonant, vowel, POS tag, goshu, accent mark and accent sandhi type
        for each mora, and used those information as the feature of $s$.

    \subsection{Neural Network Model  $f$}\label{sec:network}
        \begin{figure}[!t]
            \begin{minipage}[t]{1.00\linewidth}
                  \centering
                  \centerline{\includegraphics[width=1.\columnwidth]{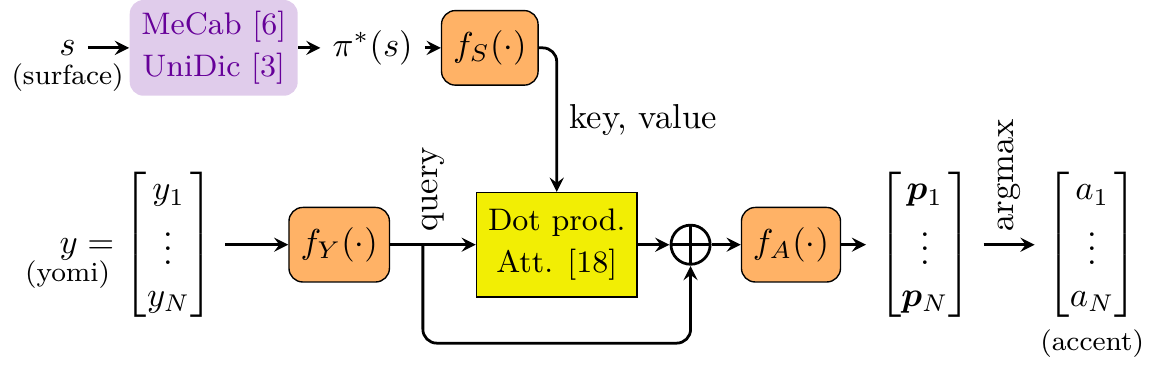}}
                  \vspace{-5pt}
                  \caption{
                    Network structure.
                  }\label{fig:network}
            \end{minipage}
        \end{figure}

\begin{table*}[t!]
    \begin{minipage}[t]{0.95\linewidth}
        \caption{Comparison of UniDic and our dictionary based on NEologd. Wavy lines indicate errors.}
        \label{tab:samplesent}
        \footnotesize{
            \centering
            \begin{tabular}{@{}p{0.23\columnwidth}p{0.08\columnwidth}p{0.65\columnwidth}@{}} \hline
                input text & dictionary & simplified output of MeCab
                \\ \hline
                \multirow{2}{*}{\parbox{0.23\columnwidth}{\jpn{江戸川や多摩川、荒川、隅田川、神田川などがある。}}}
                & UniDic
                &
                \uwave{e\age{}do\vsp{}ka\age{}wa\sage{}}\vsp{}ya\vsp{}\uwave{ta\sage{}ma\vsp{}ka\age{}wa\sage{}}\vsp{},
                \vsp{}a\age{}rakawa\vsp{},
                \vsp{}\uwave{su\age{}mida\vsp{}ka\age{}wa\sage{}}\vsp{},
                \vsp{}\uwave{ka\age{}\NN{}da\vsp{}ka\age{}wa\sage{}}\vsp{}nado\vsp{}ga\vsp{}a\sage{}ru\vsp{}.
                \\
                & ours
                &
                e\age{}dogawa\vsp{}ya\vsp{}ta\age{}ma\sage{}gawa\vsp{},\vsp{}a\age{}rakawa\vsp{},
                \vsp{}su\age{}mida\sage{}gawa\vsp{},\vsp{}ka\age{}\NN{}da\sage{}gawa\vsp{}nado\vsp{}ga\vsp{}a\sage{}ru\vsp{}.
                \\ \hline
                \multirow{2}{*}{\parbox{0.23\columnwidth}{\jpn{浦島太郎が竜宮城でもらった玉手箱を開けると、}}}
                & UniDic
                &
                u\age{}ra\sage{}shima\vsp{}ta\sage{}roo\vsp{}ga\vsp{}
                \uwave{ryu\sage{}u\vsp{}mi\sage{}yagi}\vsp{}de
                \vsp{}mo\age{}ra\QQ{}\vsp{}ta\vsp{}ta\age{}mate\vsp{}ba\age{}ko\vsp{}o\vsp{}a\age{}keru\vsp{}to\vsp{},
                \\
                & ours
                &
                u\age{}ra\sage{}shima\age{}ta\sage{}roo\vsp{}ga\vsp{}ryu\age{}uguujoo\vsp{}de\vsp{}mo\age{}ra\QQ{}\vsp{}ta\vsp{}ta\age{}mate\sage{}bako o\vsp{}a\age{}keru\vsp{}to\vsp{},
                \\ \hline
                \multirow{2}{*}{\parbox{0.23\columnwidth}{\jpn{kubernetes と docker と nginx の使い方を覚える。}}}
                & UniDic
                &
                \uwave{\textless{}UNK\textgreater{}}\vsp{}to\vsp{}
                \uwave{\textless{}UNK\textgreater{}\vsp{}}to\vsp{}
                \uwave{e\sage{}nu\vsp{}ji\sage{}i\vsp{}a\sage{}i\vsp{}e\sage{}nu\vsp{}e\sage{}\QQ{}kusu}\vsp{}
                no\vsp{}tsu\age{}kai\vsp{}kata\vsp{}o\vsp{}o\age{}boe\sage{}ru\vsp{}.
                \\
                & ours
                &
                ku\age{}ube\sage{}netis\vsp{}to\vsp{}do\sage{}\QQ{}kaa\vsp{}to\vsp{}
                e\age{}\NN{}ji\NN{}e\sage{}\QQ{}kusu\vsp{}
                no\vsp{}tsu\age{}kai\vsp{}kata\vsp{}o\vsp{}o\age{}boe\sage{}ru\vsp{}.
                \\ \hline
                \multirow{2}{*}{\parbox{0.23\columnwidth}{\jpn{ラグビー日本代表の試合を見に飛田給に行く。}}}
                & UniDic
                &
                ra\sage{}gubii\vsp{}ni\age{}\QQ{}po\sage{}\NN{}\vsp{}da\age{}ihyoo\vsp{}no\vsp{}shi\age{}ai\vsp{}o\vsp{}mi\sage{}\vsp{}ni\vsp{}
                \uwave{hi\age{}da\vsp{}kyuu}
                \vsp{}ni\vsp{}
                i\age{}ku\vsp{}.
                \\
                & ours
                &
                \uwave{ra\age{}gubiiniho\NN{}da\sage{}ihyoo}\vsp{}no\vsp{}
                shi\age{}ai\vsp{}o\vsp{}mi\sage{}\vsp{}ni\vsp{}to\age{}bita\sage{}kyuu\vsp{}ni\vsp{}i\age{}ku\vsp{}.
                \\ \hline
            \end{tabular}
        }
    \end{minipage}
\end{table*}

        We used a simple neural network model shown in
        Fig.~\ref{fig:network}.
        The network includes three trainable submodules $f_S(\cdot), f_Y(\cdot)$, and $f_A(\cdot)$.
        $f_S(\cdot)$ and $f_Y(\cdot)$ encode the surface $\pi^*(s)$ and the yomi $y$, respectively.
        Then the dot-product attention~\cite{dotprodatt} aligns them,
        and finally, $f_A(\cdot)$ decodes it and outputs the accent $\bm{a}$.

        The main body of each $f_\bullet(\cdot)$ was a four-layer non-causal 1D convolutional network.
        Additionally, $f_S(\cdot)$ and $f_Y(\cdot)$ were preceded by the character-embedding layers,
        and $f_A(\cdot)$ was followed by point-wise layers.
        Each convolutional layer of $f_\bullet(\cdot)$-s
        was a 1D dilated convolution of kernel size $3$ and channel size $64$,
        preceded by a dropout ($p=0.5$), followed by a batch renormalization~\cite{ioffe2017batch} and
        a highway activation~\cite{highway} (gated residual connection).
        The dilation factors of the convolutions of $f_\bullet(\cdot)$ were $1\to3\to1\to3$.

        The objective function was the cross entropy between the predicted density
        $\bm{p}_t$ and the smoothed ground truth~\cite{labelsmoothing1,labelsmoothing}.
        (We intentionally gave the wrong label with a probability of 60\%,
        while the correct label with a probability of 40 \%,
        to prevent our model to be overconfident.)
        We also added another loss function on attention matrix~\cite{tachi},
        which promotes the attention matrix to be diagonal.

    \subsection{Training Data}
        To train the above model, we annotated a portion of the words in NEologd.
        We first roughly classified the words of NEologd
        as shown in Table~\ref{tab:neologdwords} using simple regular expressions,
        excluding some noisy words, such as kanji words written in katakana\footnote{
            For example, we ignored $s=$\jpn{トウキョウトチジセンキョ}, which is very unnatural.
            It is almost always written in kanji as \jpn{東京都知事選挙}.
        }.
        The classification was not perfect and we found many misclassifications,
        but we did not modify them as it was infeasible to correct them manually.

        Then, for each category, we sampled the words, the number of which is shown in Table~\ref{tab:neologdwords}~(b).
        Then the first author of this article, a native speaker of contemporary Tokyo dialect, annotated those words.
        The author did not know the exact accents of the most of the extracted words,
        but entered plausible ones that would sound natural.
        Some of the yomis of NEologd were wrong,
        but the author entered plausible accents assuming that these yomis are correct.

        In addition to these data,
        we used 500 sentences,
        7,000 UniDic words, and
        20,000 synthetic compound words.
        To synthesize those compound words, we randomly sampled nouns
        from UniDic, and concatenated them by either of following two rules.
        \begin{align*}
            (1)~ &s = s_1 | s_2, ~~~~ y = y_1 | \eta | y_2 \\
            (2)~ &s = s_1 | s_2 | \xi | s_3 | s_4, ~~~~ y = y_1 | \eta_1 | y_2 | \xi | y_3 | \eta_2 | y_4
        \end{align*}
        where `$|$' denotes the string concatenation,
        $\eta, \xi$ are randomly drawn from $\eta \in \{``", \mathit{no}, \mathit{ga}, \mathit{tsu}, \mathit{wa}\}$,
        and $\xi \in \{\textit{to}, \textit{wa}, \textit{ga}, \textit{no}, \textit{mo}\}$, respectively.
        In either case, we defined the accent using Sagisaka's rule~\cite{sagisaka}.
        We thus obtained nonsense compound words e.g.
        \jpn{海浜しめじ茸と炭酸兎} ({ka\age{}ihi\NN{}shimeji\sage{}taketota\age{}\NN{}sa\NN{}gau\sage{}sagi}).

\section{Experiment}

\begin{table}[t]
    \begin{minipage}[t]{0.99\linewidth}
        \caption{
            Examples of correctly estimated accents.
            None of these words are included in the training data.
            Other examples are also found in Table~\ref{tab:samplesent}.
            (The mark \starsimple{} indicates that the top result of MeCab+UniDic analysis of the word is incorrect.)
        }\label{tab:estimated_correct}%
        \vspace{-15pt}%
        \begin{center}%
            {\scriptsize%
                \fbox{
                \begin{tabular}{@{}p{0.99\columnwidth}@{}}
                        {\jpn{機械学習}} (ki\age{}kaiga\sage{}kushuu),
                        {\jpn{清涼飲料水}} (se\age{}iryooi\NN{}ryo\sage{}osui),
                        {\jpn{量子コンピューター}} (ryo\age{}oshiko\NN{}pyu\sage{}utaa),
                        {\jpn{リチウムイオン電池}} (ri\age{}chiumuio\NN{}de\sage{}\NN{}chi),
                        {\jpn{モバイルバッテリー}} (mo\age{}bairuba\sage{}\QQ{}terii),
                        {\jpn{京都タワー}} (kyo\age{}otota\sage{}waa),
                        {\jpn{五稜郭}} (go\age{}ryo\sage{}okaku),
                        {\jpn{横浜赤レンガ倉庫}} (yo\age{}kohamaakare\NN{}gaso\sage{}oko),
                        {\jpn{江戸東京博物館}} (e\age{}dotookyoohakubutsu\sage{}ka\NN{}),
                        {\jpn{御御御付け\starsimple}} (o\age{}mio\sage{}tsuke),
                        {\jpn{36協定\starsimple}} (sa\age{}burokukyo\sage{}otee),
                        {\jpn{八ッ場ダム\starsimple}} (ya\age{}\NN{}bada\sage{}mu),
                        {\jpn{井戸端会議\starsimple}} (i\age{}dobataka\sage{}igi),
                        {\jpn{赤血球\starsimple}} (se\age{}\QQ{}ke\sage{}\QQ{}kyuu),
                        {\jpn{黄色ブドウ球菌\starsimple}} (o\age{}oshokubudookyu\sage{}uki\NN{}),
                        {\jpn{Python\starsimple}} (pa\sage{}iso\NN{}),
                        {\jpn{word2vec\starsimple}} (wa\age{}adotsuube\sage{}\QQ{}ku),
                        {\jpn{Led Zeppelin\starsimple}} (re\age{}\QQ{}dotse\sage{}\QQ{}peri\NN{}),
                        {\jpn{FreeBSD\starsimple}} (fu\age{}riibiiesudi\sage{}i),
                        {980hPa\starsimple} (kyu\sage{}uhyakuha\age{}chijuuhekutopa\sage{}sukaru),
                        {\jpn{2468円}} (ni\age{}se\sage{}\NN{}\age{}yo\sage{}\NN{}hyakuro\age{}kujuuhachi\sage{}e\NN{}),
                        {\jpn{W杯\starsimple}} (wa\age{}arudoka\sage{}\QQ{}pu),
                        {\jpn{九蓮宝燈\starsimple}} (chu\age{}ure\NN{}po\sage{}otoo),
                        {\jpn{平昌オリンピック\starsimple}} (pyo\age{}\NN{}cha\NN{}ori\NN{}pi\sage{}\QQ{}ku),
                        {\jpn{棒々鶏\starsimple}} (ba\age{}\NN{}ba\sage{}\NN{}jii),
                        {\jpn{東京都国立市\starsimple}} (to\age{}okyo\sage{}otoku\age{}nitachi\sage{}shi),
                        {\jpn{目黒のさんま}} (me\sage{}guronosa\age{}\NN{}ma),
                        {\jpn{東海道五十三次\starsimple}} (to\age{}oka\sage{}idoogo\age{}juusa\sage{}\NN{}tsugi),
                        {\jpn{世界の終わりとハードボイルドワンダーランド}} (se\sage{}kainoowarito
                        ha\age{}adoboirudowa\NN{}daara\sage{}\NN{}do),
                        {\jpn{東京都道・埼玉県道25号飯田橋石神井新座線}} (to\age{}okyooto\sage{}doo%
                        sa\age{}itamake\sage{}\NN{}doo%
                        \age{}ni\sage{}juu\age{}go\sage{}goo%
                        i\age{}ida\sage{}bashi%
                        sha\age{}kuji\sage{}i%
                        ni\age{}izase\NN{})
                    \\
                \end{tabular}
                }
            }
        \end{center}
    \end{minipage}
\end{table}

\begin{table}[t]
    \begin{minipage}[t]{0.99\linewidth}
        \caption{
            Examples of errors
        }\label{tab:estimated_wrong}%
        \begin{center}%
            {\scriptsize%
            \begin{tabular}{@{}lll@{}}
                \hline
                surface & estimated accent & the author's accent \\ \hline
                \jpn{大学院} & \xmarksimple{} da\age{}igakui\NN{} & da\age{}igaku\sage{}i\NN{} \\
                \jpn{信号処理} & \xmarksimple{} shi\age{}\NN{}go\sage{}oshori & shi\age{}\NN{}goosho\sage{}ri \\
                \jpn{ケンタウルス座} & \xmarksimple{} ke\age{}\NN{}taurusu\sage{}za & ke\age{}\NN{}taurusuza \\
                \jpn{ドラム式洗濯機} & \xmarksimple{} do\age{}ramu\sage{}shikise\NN{}ta\sage{}ku\sage{}ki & do\age{}ramushikise\NN{}ta\sage{}kuki \\
                \jpn{明治神宮前} & \xmarksimple{} me\sage{}ijiji\age{}\NN{}guuma\sage{}e & me\age{}ijiji\NN{}guuma\sage{}e \\
                \jpn{小竹向原} & \xmarksimple{} ko\sage{}takemu\age{}ka\sage{}ihara & ko\age{}takemukai\sage{}hara \\
                \jpn{武蔵小杉} & \xmarksimple{} mu\sage{}sashiko\age{}sugi & mu\age{}sashiko\sage{}sugi \\
                \jpn{東京都渋谷区} & \xmarksimple{} to\age{}okyootoshi\age{}buya\sage{}ku & to\age{}okyo\sage{}otoshi\age{}buya\sage{}ku \\
                \jpn{紅白歌合戦} & \xmarksimple{} ko\age{}ohakuutaga\sage{}\QQ{}se\NN{} & ko\sage{}ohakuu\age{}taga\sage{}\QQ{}se\NN{} \\
                \jpn{展覧会の絵} & \xmarksimple{} te\age{}\NN{}ra\NN{}ka\sage{}inoe & te\age{}\NN{}ra\sage{}\NN{}kainoe\sage{} \\
                \jpn{富嶽三十六景} & \xmarksimple{} fu\sage{}gaku\age{}sa\sage{}\NN{}juuro\age{}\QQ{}ke\sage{}i & fu\sage{}gaku\age{}sa\sage{}\NN{}juu\age{}ro\sage{}\QQ{}kei \\
                $\spadesuit$ & \xmarksimple{} su\age{}pe\sage{}edo & su\age{}peedo \\
                \jpn{(((o(*ﾟ▽ﾟ*)o)))} & \xmarksimple{} e\age{}gao & e\sage{}gao \\
                \hline
            \end{tabular}
            }
        \end{center}
    \end{minipage}
\end{table}

    \subsection{Accent Estimation Experiment}
        The experimental setting was as follows.
        We used 80\% of the annotated words for training,
        and remaining 20\% for evaluation.
        We used the Adam optimizer~\cite{adam} to train our model;
        the parameters were $(\alpha, \beta_1, \beta_2, \varepsilon) = (2\times10^{-4}, 0.5, 0.9, 10^{-5})$.
        We applied weight decay of factor $10^{-6}$ (L1 and L2 regularization) after each iteration.
        The size of mini-batch was 32.
        We trained our model for 4 days (2.5M steps).
        The version of UniDic we used was
        \textit{unidic-\allowbreak{}mecab\_\allowbreak{}kana-\allowbreak{}accent-\allowbreak{}2.1.2}\footnote{
            \url{https://unidic.ninjal.ac.jp/back_number}.~760k words.
        }.

        The evaluation criteria were as follows: the exact matching rate (EMR; the rate of the words whose estimated accents exactly matched the ground truths),
        the average hamming distance (AHD) from the ground truths,
        the precision $\mathrm{TP} / (\mathrm{TP}+\mathrm{FP})$
        and the recall $\mathrm{TP} / (\mathrm{TP}+\mathrm{FN})$
        of raise ``\age{}'' and lower ``\sage{}''.
        Table~\ref{fig:jpnwords}(c) shows the results.
        Note, considering that some words have several acceptable accents\footnote{For example,
            `ju\age{}ugo\sage{}fu\NN{}', `ju\sage{}u\age{}go\sage{}fu\NN{}', `ju\sage{}ugo\sage{}fu\NN{}' and `ju\sage{}ugofu\NN{}'
            would all be acceptable pronunciations of the word \jpn{十五分}.},
        the actual performance would be a little better than the digits shown in the Table.

        From these digits, we can say the following for most categories of the words.
        \begin{itemize}
            \item The proposed method estimated the exact accents of over a half of the words (EMR $>$ 50 \%).
            \item The number of estimation errors in a word is less than 1 on average (AHD $<$ 1).
            \item We may trust more than 80\% of ``\age{}'' prec $> 0.8$), and 75\% of ``\sage{}'' (prec $> 0.75$).
        \end{itemize}
        Table~\ref{tab:estimated_correct} shows the examples of correctly estimated accents.
        Even when UniDic provided little useful phonetic information about the words (e.g., Python, word2vec, \textit{Yanbadamu},)
        the proposed method could estimate the accent correctly using the yomis.
        On the other hand, Table~\ref{tab:estimated_wrong} shows the examples of errors.
        Note, it is sometimes impossible in principle to estimate the accent of some words
        without taking into account cultural backgrounds or customs. 
        For example,
        the accents of some place names are customary and difficult to predict even for native speakers
        unless they are familiar with the neighbourhood.
        Some of the errors may be of this kind, e.g.{} \textit{Meiji Jing\^umae}, \textit{Kotake Mukaihara}.
        The estimated accents of these words are possible grammatically, but may sound a little unnatural for local residents.

    \subsection{Application to Japanese TTS}

        We estimated the accents of all words listed in NEologd using the proposed technique,
        and obtained a new dictionary\footnote{
            We additionally modified the unigram cost of each word a little,
            because NEologd's unigram costs of some categories of words (e.g.{} person's name) were too small, in the current version.
        }.
        Table~\ref{tab:samplesent} shows some examples of the text analysis based on each dictionary.
        We also checked the effectiveness of the dictionary by using it in a TTS system.
        In our experiment, we used the system based on~\cite{tachi}.
        The input data of the system was the yomi, accent marks and POS tags.
        The training data was JSUT corpus~\cite{jsut}\footnote{
            Female voice, \# speaker is 1, approx 10 hours. We resampled all the data from 48kHz to 24kHz.
        }.
        When using UniDic, we applied the subroutine shown in footnote~\ref{footnote:numeral},
        while we did not when using our dictionary.

        Seven native speakers evaluated 40 synthesized speech signals (2 dictionaries $\times$ 20 sentences).
        Of these 20 cases, our dictionary was clearly better in 10 cases,
        UniDic was clearly better in 2 cases, and both were almost evenly evaluated in the remaining 8 cases.
        Qualitatively, we found our dictionary received lower evaluations in following cases.
        (1) Even though the estimated accent was correct,
            the neural TTS system sometimes could not synthesize the word correctly,
            especially when the word is long (e.g.{} address, street, numerals, etc.),
            or the accent pattern is complicated (e.g. more than two accent nuclei ``\sage{}'').
            This is possibly due to the mismatch between the training data and the test sentences of TTS.
            Indeed, those words were rarely used in JSUT corpus.
        (2) Some research participants did not know the yomis of some difficult words.

\section{Concluding Remarks}
    In this paper, we proposed a neural network-based
    technique to estimate the accents of Japanese words,
    using their surfaces and the yomis (phonetic information except the accent).
    The author annotated 17200 words out of 3 million words listed in NEologd,
    and trained the model.
    Experiments showed that the method estimated
    the accent of some categories of words (e.g. numerals, address, katakana words, etc.) with high accuracies,
    while the performance was not necessarily satisfactory for other categories (emoji, etc.).

    By applying it to the words of NEologd,
    we obtained a large scale accent dictionary.
    In principle, the text tokenization performance of the dictionary is as good as NEologd.
    We may expect that it can reduce Japanese TTS users' frustrations
    to modify yomis and accents of compound words, persons' names, place names, neologisms, etc.
    Although the dictionary is not complete, it can be a seed for further improvements.
    We may grow the dictionary iteratively
    through the operation of a real-world TTS system based on this baseline dictionary.
    The authors are planning to release the code of the proposed method shortly.

\bibliographystyle{IEEEbib}
{
    \bibliography{mybib}
}

\newpage
\section*{Supplementary Material}
Update Sep., 2020.
\subsection*{Trained Model and Source Code}
\begin{minipage}{2\linewidth}
    The trained model and the inference code (automatic dictionary generator) are available at the following site.
    \begin{quote}
        \small{\url{https://github.com/PKSHATechnology-Research/tdmelodic}}
    \end{quote}
\end{minipage}

\end{document}